\documentclass{article}

\usepackage[main,preprint]{neurips_2026}

\usepackage[utf8]{inputenc}
\usepackage[T1]{fontenc}
\usepackage{url}
\usepackage{booktabs}
\usepackage{amsfonts}
\usepackage{amsmath}
\usepackage{amssymb}
\usepackage{nicefrac}
\usepackage{microtype}
\usepackage{xcolor}
\usepackage{graphicx}
\usepackage{subcaption}
\usepackage{multirow}
\usepackage{makecell}
\usepackage{colortbl}
\usepackage{siunitx}
\usepackage{pifont}      
\usepackage{xspace}
\usepackage{hyperref}
\usepackage{cleveref}
\usepackage{placeins}

\usepackage{amsthm}

\definecolor{lightgray}{gray}{0.92}

\newcommand{\ahill}{\alpha_{\text{Hill}}}

\newcommand{\llada}{LLaDA-8B\xspace}
\newcommand{\dream}{DREAM-7B\xspace}
\newcommand{\llama}{Llama-3.1-8B\xspace}
\newcommand{\qwen}{Qwen-2.5-7B\xspace}

\newcommand{\lladas}{LLaDA\xspace}
\newcommand{\dreams}{DREAM\xspace}
\newcommand{\llamas}{Llama\xspace}
\newcommand{\qwens}{Qwen\xspace}

\newcommand{\pyar}{AR-160M\xspace}
\newcommand{\pydlm}{DLM-160M\xspace}

\usepackage[normalem]{ulem}
\renewcommand{\paragraph}{\textbf}
\usepackage{enumitem}
\setlist{nosep}

\title{Layer Collapse in Diffusion Language Models}

\author{%
Alexander Conzelmann$^{1,2}$ \quad Albert Catalan-Tatjer$^{1,2,3}$ \quad Shiwei Liu$^{1,2,3}$ \\
$^1$Tübingen AI Center \\ 
$^2$Max Planck Institute for Intelligent Systems \\
$^3$ELLIS Institute Tübingen \\
Correspondence to: \texttt{a.conzelmann@uni-tuebingen.de} 
}

\begin{document}

\maketitle
 
\begin{abstract}
Diffusion language models (DLMs) have recently emerged as competitive alternatives to autoregressive (AR) language models, yet the differences in their activation dynamics remain poorly understood.
We characterize these dynamics in LLaDA-8B and identify a striking layer-collapse property: a few early layers exhibit highly similar, collapsed activation patterns dominated by a single large super-outlier that persists over a long token range.
Despite its apparent redundancy, we have found that this outlier is critical: pruning it causes the model's outputs to degrade into repetitive random token loops.
Paradoxically, besides this outlier, layers in LLaDA contain more redundant representations, where redundancy is particularly pronounced in earlier layers. This pattern is the reverse of what is commonly observed in AR language models, where deeper layers tend to become more redundant due to undertraining, as measured by representation similarity. Our analysis further indicates that layer collapse in DLMs is not driven by undertraining. Rather, it appears to arise from overtraining: a dominant outlier becomes an indispensable carrier of information, while the remaining representations collapse into redundant structure. These observations have strong practical implications, as we verify through controlled pre-training experiments.
First, DLMs are surprisingly robust to compression: performance of LLaDA under 3-bit GPTQ quantization drops only by -1.8\%
on GSM8K, whereas Llama-3.1-8B under the same settings drops by -64.7\%. Second, optimal non-uniform sparsity allocation reverses between the two model families:
under an average budget of 50\% sparsity, allocating more sparsity to early layers in LLaDA yields +8.4\% over the reverse strategy, while for Llama the same early-layer-sparse allocation incurs -8.4\%. Our findings reveal that the DLM training objective fundamentally reshapes layer dynamics relative to AR models, with direct consequences for how such models should be compressed and deployed. Our code is available at \url{https://github.com/Conzel/super-outlier-dlm}.

\end{abstract}
\vspace{-12pt}
\section{Introduction}
\vspace{-6pt}
\label{sec:intro}

Diffusion language models (DLMs) have recently emerged as a competitive alternative to
autoregressive (AR) large language models. Open models such as \llada~\citep{nieLargeLanguageDiffusion2025}
and \dream~\citep{yeDream7BDiffusion2025} match the quality of AR counterparts at comparable scale
on standard reasoning and language understanding benchmarks. Rather than producing one token at a
time from left to right, DLMs denoise a masked sequence over multiple refinement steps, which
enables parallel token generation
\citep{wuFastdLLMTrainingfreeAcceleration2025,chenDParallelLearnableParallel2025,maDInferEfficientInference2025}
and avoids the reversal curse \citep{berglund2024reversalcursellmstrained}. 
As DLMs become more widespread, questions on the structure of their internal representations 
and how they propagate through layers arise. Of particular interest is whether the findings 
from AR models transfer one-to-one to DLMs or if there are particular behaviours that necessitate 
extra care during tasks such as training, fine-tuning or compression.

\begin{figure}[ht]
    \centering
        \includegraphics[width=1.0\textwidth]{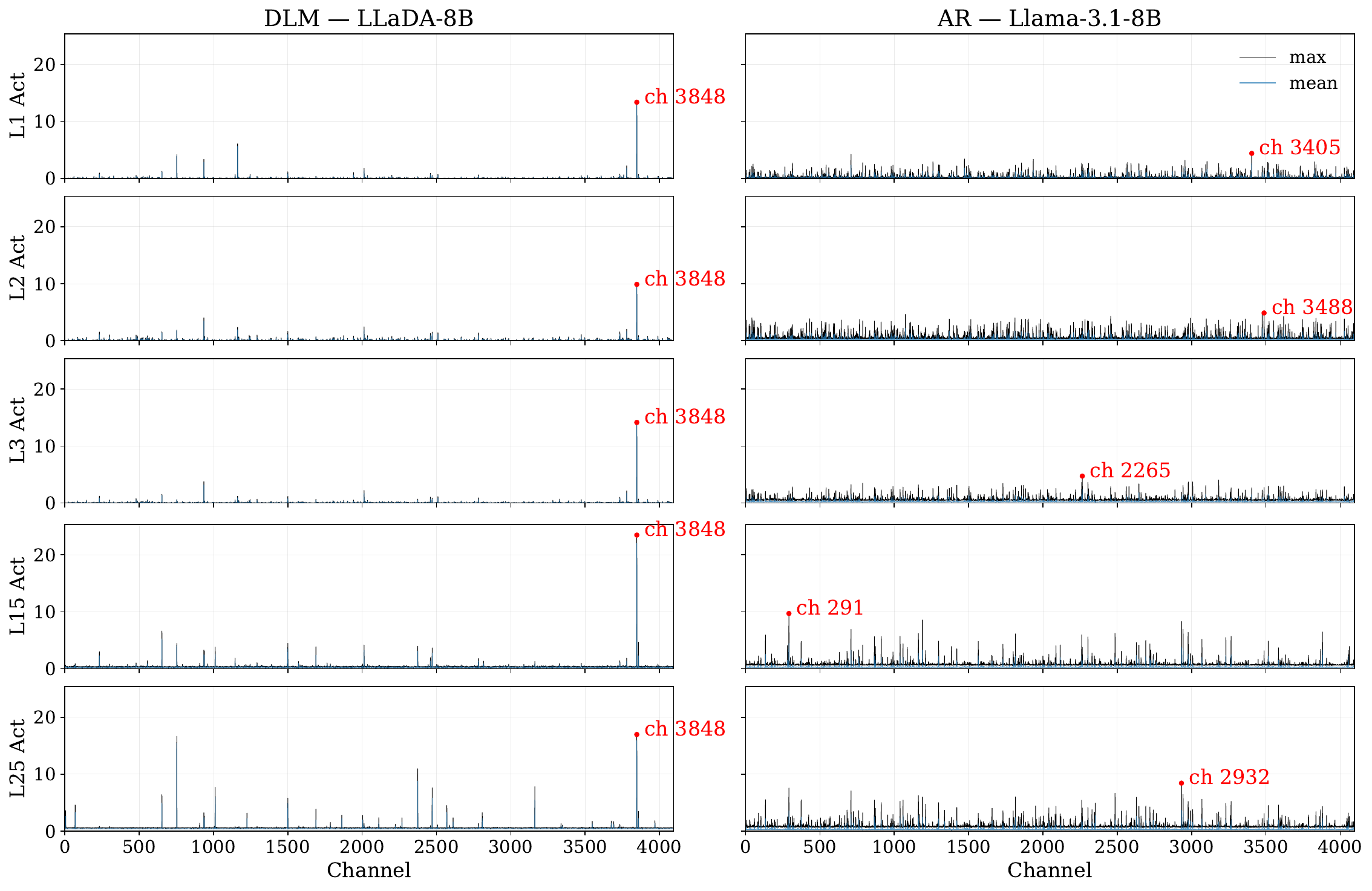}
        \caption{\lladas activations contain a single consistent, very large super-outlier channel which persists well into the middle layers. The marked dot shows the channel with the highest activation magnitude maximum (averaged over tokens and sequence positions). 
        For \llamas, the dominant channel changes on almost every layer.}\label{fig:sparkline}
\end{figure}

In this paper, we find that DLMs show distinct and partly opposite behaviour from AR models.
We find that \llada's internal representations are qualitatively unlike those
of any AR model previously studied. 
Most strikingly, a \emph{single} activation channel (see \autoref{fig:sparkline}) remains persistently and highly activated across all tokens throughout the first half of the model's layers — an extreme outlier that drives layer collapse, causing multiple layers to produce nearly identical, redundant hidden representations. 
This channel dominates the model to such a degree that ablating it alone causes a near-total collapse in capability. While outlier channels are prominent in AR models (see \citet{dettmers2022gpt3, yu2025superweightlargelanguage}), they only cause extreme activations for specific tokens or token positions, and pruning a single one of them causes only a noticeable accuracy drop (-4\% in our experiment, see \autoref{tab:reduce-gsm8k-acc}), but not complete model collapse.
The layer-wise similarity structure is also inverted (see \autoref{fig:similarity-plots}). In standard AR models, early layers usually learn more distinct transformations, while deeper layers become increasingly redundant as a consequence of undertraining \citep{sun2025curse}.
By contrast, \llada exhibits highly redundant early layers. We find that this redundancy is driven by the dominating super-outlier, which suppresses the representation expressivity of other channels across the first few layers.

At first glance, this observation conflicts with the curse-of-depth hypothesis \citep{sun2025curse}, which attributes high representation similarity in Pre-LN Transformers to the undertraining of deeper layers. However, our weight spectral density analysis shows that DLMs exhibit a compatible layerwise trainability pattern, but with a different source of redundancy. In \llada, the unusually high similarity among early layers is not caused by undertraining; instead, it arises from the relative overtraining of these layers, resulting in dominant super-outliers concentrated in the earliest layers.

\textbf{Why activation structure matters.} A long line of work has shown that algorithms
which explicitly account for activation structure
\citep{lin2023awq,lee2024owqoutlierawareweightquantization,quaRot2024} substantially outperform
naive counterparts on quantization and pruning. The structure itself has been mapped in detail
for AR models: \emph{outlier features}, extreme activations found in essentially all sufficiently
large language models \citep{dettmers2022gpt3,an2025systematicoutlierslargelanguage}, are tied to
core information-processing functions. \emph{Super-weights} suppress stop-tokens
\citep{yu2025superweightlargelanguage}, and attention outliers at sequence boundaries serve as
sinks for excess attention mass \citep{gu2024attention,xiao2024efficientstreaminglanguagemodels}.
Early DLM work hints that the picture is different: \citet{rulli2025attentionsinksdiffusionlanguage}
report that attention sinks are present in DLMs but far less sensitive to removal, an observation
that \citet{myrzakhan2026sinkawarepruningdiffusionlanguage} have already exploited for more
efficient pruning. Our results push this line considerably further, and identify the diffusion training objective itself as the source of the divergence.

\textbf{Contributions.} We close part of the knowledge gap between AR and diffusion language
models through a systematic study of activation histograms and layer similarities in \llada.
To separate effects arising from the diffusion objective from architectural quirks, we compare
\llada to \llama, which share a near-identical architecture, and complement this with the first
controlled pre-training comparison in which an AR and a DL model are trained under identical
conditions. 
\vspace{-3pt}
\begin{itemize}
  \item We identify a \emph{super-outlier} in \llada: a single dominant activation channel qualitatively distinct from the weight and activation outliers previously reported in AR LLMs, and whose removal collapses the model.
  \item We characterize the layer-similarity structure of \llada and show that it differs markedly from that of comparable AR models, with early layers being highly redundant rather than distinct, corroborated by weight spectral analysis indicative of
    significant overfitting.
  \item We translate these observations into practical guidance for DLM compression and show that AR-derived sparsity heuristics are not merely suboptimal but actively inverted relative to what \llada calls for.
  \item To isolate the role of the training objective, we pre-train an AR and a DL model under an identical setup and show that the diffusion objective is responsible for the observed layer and compression behaviours. To our knowledge, this is the first such controlled comparison.
\end{itemize}

\vspace{-6pt}
\section{Layer Dynamics in Diffusion Language Models}
\label{sec:dynamics}
\vspace{-3pt}
\subsection{Metrics}
\vspace{-3pt}
\label{sec:collapse}
\textbf{Layer Similarity.} Our main method of analysis is the per-token cosine similarity between the hidden activations of different layers of a model.
Let $h_i(x, t) \in \mathbb{R}^d$ denote the hidden state at layer $i$ for input sequence $x \sim p_x$ at token position $t$. We define the layer-wise similarity as
\begin{equation}
    \mathrm{sim}(i,j) \;=\; \mathbb{E}_{x \sim p_x} \, \mathbb{E}_{t} \left[ \frac{\langle h_i(x, t),\, h_j(x, t) \rangle}{\|h_i(x, t)\|_2 \, \|h_j(x, t)\|_2} \right],
    \label{eq:cosine}
\end{equation}
where the inner expectation is taken over all (non-padding) token positions $t$ in the sequence.
A value of $\mathrm{sim}(i,j)$ close to $1$ indicates that layers $i$ and $j$ produce nearly collinear representations, which we use as a proxy for layer redundancy: if the representation barely changes between two layers, the intermediate computation contributes little and is a natural candidate for pruning.

\textbf{Estimating Activations.}
In practice, we estimate all hidden layer activations using a fixed calibration set of 128 sequences of 
length 2048 drawn from the C4 corpus. For DLMs we duplicate each sample 4 times and mask a $t \sim \text{Unif}[0,1]$ fraction of  tokens to imitate the activation statistics of a partly decoded sample sequence. We note however that this  has very little effect on the resulting figures (see \autoref{app:no-mask}, repeating similarity and activation plots without masking), indicating that our findings are not sensitive to the  exact decoding process that the DLM implements.


\textbf{Heavy-tailedness of the Weight Spectrum as a Layer Trainability Measure.}
Representation similarity is a useful diagnostic for identifying redundant layers, but it is not by itself a reliable measure of layer trainability. High similarity can arise from qualitatively different mechanisms: in AR Transformers, similar representations in deeper layers are often associated with under-trained, near-identity residual mappings \citep{sun2025curse}, whereas in DLMs, highly similar layers may instead reflect strong training of a few dominant directions, manifested as distinguished spectral or activation outliers. Thus, similarity-based importance metrics can be misleading, since they conflate under-utilized layers with layers whose representations are highly structured but dominated by a small number of directions.

Motivated by Heavy-Tailed Self-Regularization (HT-SR) theory~\citep{martinTraditionalHeavyTailedSelf2019,luAlphaPruningUsingHeavyTailed2024,he2026one}, we argue that the heavy-tailedness of the weight spectrum provides a more reliable, training-data-free measure of layer quality than representation similarity. HT-SR theory suggests that the empirical spectral density of neural-network weight matrices encodes information about the degree of learned structure: heavier-tailed spectra are typically associated with stronger feature learning and better-trained layers. In our setting, this spectral perspective is particularly useful because it helps distinguish two cases that representation similarity alone cannot separate: near-identity redundancy caused by insufficient training, and apparent redundancy caused by strong low-dimensional structure or outlier-dominated spectra.

To quantify the heavy-tailedness of a layer, we use the Hill estimator $\ahill$~\citep{luAlphaPruningUsingHeavyTailed2024}. For weight $W$, let $\lambda_1 \geq \lambda_2 \geq \dots \geq \lambda_N$ denote the eigenvalues of $W^\top W$ sorted in decreasing order. The Hill estimator of the power-law tail exponent, computed from the top $k$ eigenvalues, is
\vspace{-4pt}
\begin{equation}
    \ahill(k) \;=\; 1 + \frac{k}{\sum_{i=1}^{k} \log \frac{\lambda_{n-i+1}}{\lambda_{n-k}}},
    \label{eq:alpha-hill}
\end{equation}
where smaller values of $\ahill$ indicate heavier-tailed spectra. We set $k$ using the fix-finger method \citep{yang2023fixfinger}. Under HT-SR theory, a heavier tail is typically interpreted as evidence of stronger feature learning. 


\vspace{-6pt}
\subsection{Super-Outlier in \llada}\label{sec:super-outlier}
\begin{figure}[ht]
\vspace{-10pt}
    \centering
    \begin{subfigure}[t]{0.49\textwidth}
        \centering
        \includegraphics[width=\textwidth]{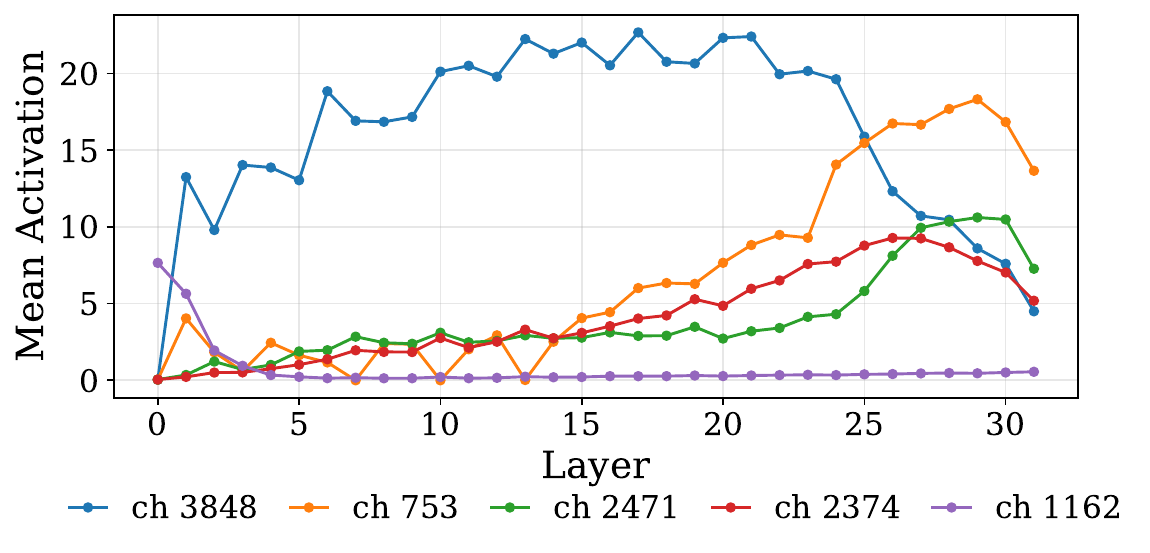}
        \vspace{-16pt}
        \caption{\llada: the super-outlier channel dominates the top-5 magnitudes consistently across early-middle layers.}\label{fig:llada-top5}
    \end{subfigure}
    \hfill
    \begin{subfigure}[t]{0.49\textwidth}
        \centering
        \includegraphics[width=\textwidth]{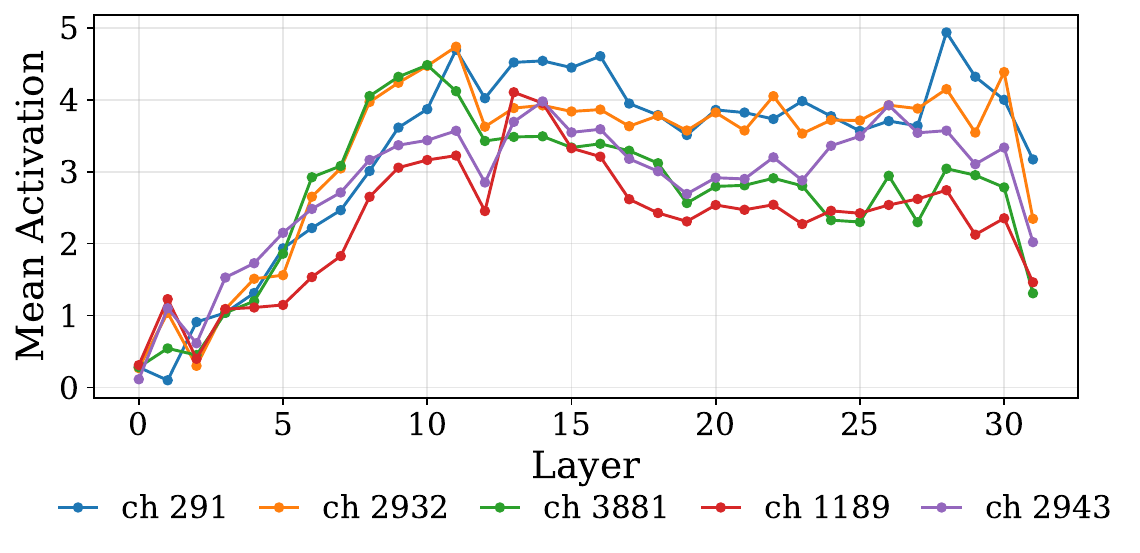}
        \vspace{-16pt}
        \caption{\llama: top-5 channels are of comparable magnitude; dominant channel changes across layers.}\label{fig:llama-top5}
    \end{subfigure}
    \vspace{10pt}
    \caption{Magnitudes of the top-5 QKV input channels across layers. In \llada, one channel persistently dominates by a wide margin, whereas in \llama no single channel is dominant.}\label{fig:top5-pair}
\end{figure}

\vspace{-6pt}
Activation outliers are a well observed phenomenon in LLMs, but those spikes usually only occur briefly for specific tokens 
or for specific token positions \citep{an2025systematicoutlierslargelanguage}. 
In the case of \llada, we present a unique finding: when analyzing the activation heatmaps over sequence positions in \lladas, we find
that there exists one single dominant outlier channel, which has a persistently high activation (up to 5 times as large as the next largest outlier)
over all sequence positions (see \autoref{fig:top5-pair}).

In contrast to outliers in AR models, which are reported to have well behaved mechanisms such as stop-word suppression \citep{yu2025superweightlargelanguage}, 
the super-outlier in \lladas seems more akin to a constant bias that is learned by the network.
The qualitative effect of pruning the super-outlier is striking: \llada degenerates into repetitive token loops, regardless of the prompt. As an illustrative example (taken from GSM8K):
\vspace{-6pt}
\begin{quote}
    \textbf{Question.} Kylar went to the store to buy glasses for his new apartment. One glass costs \$5, but every second glass costs only 60\% of the price. Kylar wants to buy 16 glasses. How much does he need to pay for them?

    \textbf{\llada answer (super-outlier pruned).} \textit{buy buy buy buy1yl buy buy buy}
\end{quote}
\vspace{-6pt}
This is in stark contrast to \llama, where removing the highest magnitude channel only slightly reduces the accuracy, see \autoref{tab:reduce-gsm8k-acc}.
\begin{table}[ht]
    \centering
    \vspace{-10pt}
    \caption{
    Effect of pruning the single highest-magnitude channel on GSM8K (small subset). Removing channel 3848 in \llada (the super-outlier) collapses the model entirely, while pruning the corresponding top-magnitude channel in \llama causes only a minor accuracy drop.}
    \begin{tabular}{lccc}
        \toprule
        Model & Channel pruned & Accuracy & $\Delta$ baseline \\
        \midrule
        \llama  & 291  & 79\% & $-4\%$ \\
        \llada  & 3848 & 0\%  & $-83\%$ \\
        \bottomrule
    \end{tabular}
    \label{tab:reduce-gsm8k-acc}
\end{table}

\subsection{Early-Layer Redundancies in \llada }
\begin{figure}
    \begin{center}
        \includegraphics[width=0.95\textwidth]{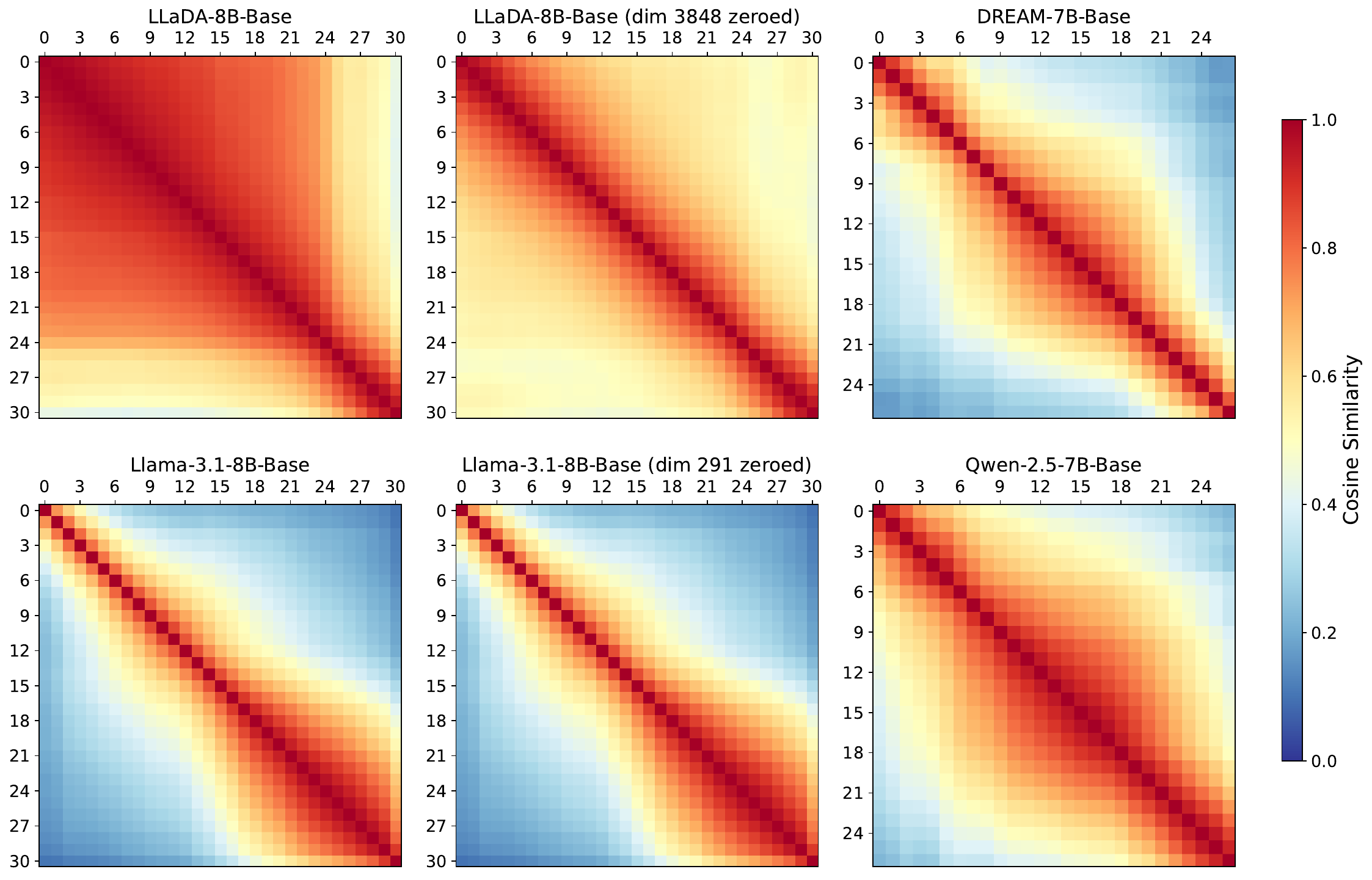}
    \end{center}
    \caption{Per-token cosine similarity for different models. The top row shows DLMs, the bottom shows ARs. The left column shows a comparison of \llada 
        and \llama, where \llada exhibits highly redundant (collapsed) layer similarities.
        For the middle column, we have removed the largest outlier from the similarity calculations, showing  
        that much of the layer similarity of \lladas is due to a single outlier, unlike \llamas, where removing one outlier barely changes the resulting similarities. 
        The right column shows a comparison between \qwen and \dream.
        As \dreams was finetuned from \qwens-initialized weights, their layer similarity patterns are very similar. }\label{fig:similarity-plots}
    \vspace{-6pt}
\end{figure}

Next, we analyze the pairwise similarities of the hidden representations in \lladas and \llamas in \autoref{fig:similarity-plots}.
Surprisingly, we find that almost all layers in \lladas (top left) exhibit an extremely high similarity to each other, 
especially early layers which are almost identical to each other even over ranges of 15+ layers. 
This is partially mitigated if the similarities are calculated with the channel corresponding to the super-outlier set to zero (Ch. 3848, see \autoref{sec:super-outlier}), 
but the learned representations for \lladas are still more redundant than the ones from \llama (which will result in a stronger robustness of
DLMs compared to ARs, see \autoref{sec:experiments}).
On the other hand, for \llama (bottom left), the early layers show very distinct activation patterns, and the later layers contain more redundancy, 
in line with previous work on AR models.

This behaviour from \llada is not shared with \dream, since DREAM has been initialized from the weights of \qwen (an AR model) and only fine-tuned
on masked sequences. Therefore, the final weights of \dreams do not differ much from the ones of \qwens, and their similarity patterns look almost identical.  This outlier profile can be harmful in handling DLMs. Common wisdom is that strong outliers make for layers or networks
that are hard to compress; thus the compression community has devised various ways of dealing with them 
\citep{lin2023awq,dettmers2022gpt3,quaRot2024}. 
Based on this, metrics such as OWL \citep{yinOutlierWeighedLayerwise2025} allocate a lower sparsity level to layers with strong outliers 
(which are presumably difficult to compress). In \llada, this would mean that early layers should be pruned less aggressively. Due to the redundancy that is exhibited in the early layers, we have a different finding:
early layers can actually be pruned more strongly; we quantify this in \autoref{sec:experiments}.

\vspace{-6pt}
\FloatBarrier
\subsection{Heavy-Tailed Evidence of Overtraining}
\vspace{-6pt}
\label{sec:alphahill}
Leveraging the analysis performed by~\citet{martinTraditionalHeavyTailedSelf2019,luAlphaPruningUsingHeavyTailed2024}, 
we can also measure how well trained the layers in a neural network are via the $\ahill$ estimator. 

According to the works of \citep{zhou2023temperature,liuBalancing2024, he2025alphadecaymodulewiseweightdecay}, a balanced distribution of $\ahill$ indicates, that the layers in a network a better trained. 
A relatively low value of $\ahill$ (with respect to the other layers) indicates a relatively ''overtrained'' layer, while a very high value of $\ahill$ indicates an ''undertrained'' layer.

\begin{figure}[ht]
    \begin{center}
        \includegraphics[width=0.8\textwidth]{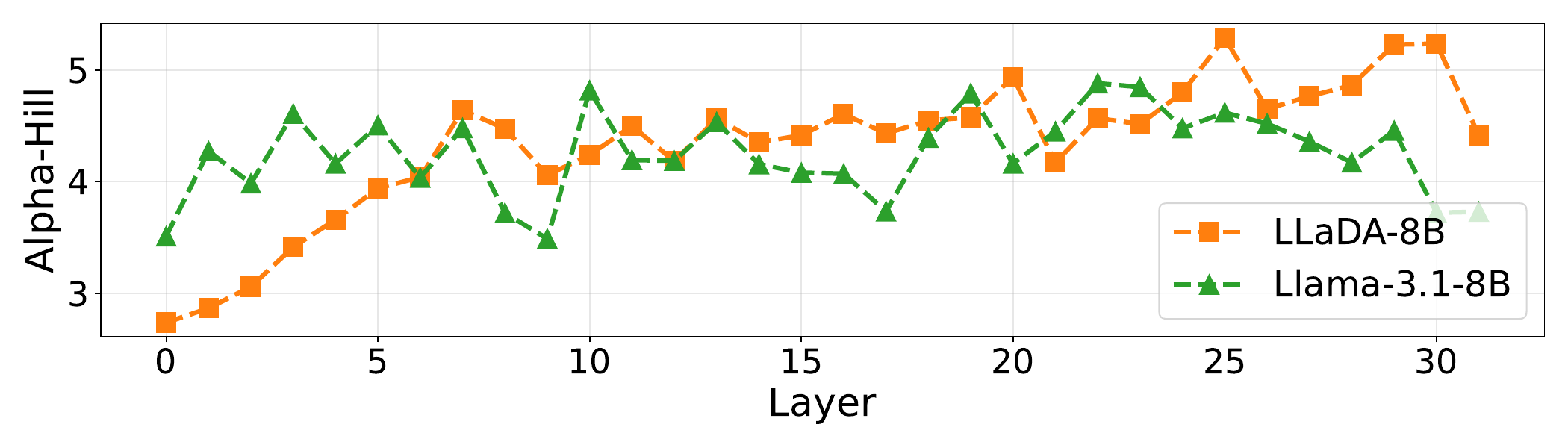}
    \end{center}
    \caption{The $\ahill$ value is very low in the early layers of \llada which shows that these layers 
    are actually overtrained. $\ahill$ is averaged over modules in layer.}\label{fig:alpha-hill-combined}
\end{figure}

This behaviour can be seen in \autoref{fig:alpha-hill-combined}: early layers in \llada have a strikingly 
low $\ahill$ (relative) value, while the distribution of the twin-model \llama is much more flat. This ties in with the early-layer occurrence of the super-outlier, a form of 
over-training where the layer representations collapse to a single
dominant channel.

\vspace{-6pt}
\section{DLMs Are More Robust to Compression Despite Super-Outliers}
\label{sec:experiments}
\vspace{-6pt}
In this section, we demonstrate that our observations have practically relevant consequences when dealing with DLMs. For this, we have chosen 
the task of model compression.  We prune and quantize \lladas and \llama using GPTQ and WANDA and measure their performance on math and question-answering under different overall and layer-wise compression strength allocations.
We show that (a) AR-optimal schemes for layer-wise compression strength are inverted when dealing with DLMs and (b) DLMs have more compression robustness. 
Observation (a) is a direct consequence of our layer-wise similarity analysis, but observation (b) is particularly surprising, as \llada contains a super-outlier that is very sensitive to change. However, 
we have seen in \autoref{fig:sparkline} that \lladas contains fewer outliers apart from the super-outlier, and we have seen in 
\autoref{fig:similarity-plots} that \lladas layer activations are more redundant, even when the outlier is removed. This translates to an overall 
stronger compression robustness, particularly as modern compression methods (like WANDA) are designed to be robust to a certain amount of outliers.

\vspace{-6pt}
\subsection{Experiment Setup}
\vspace{-6pt}
\label{sec:experiment-setup}
\paragraph{Evaluation.} Models are evaluated both on
an average of 6 common question-answering (QnA) tasks (ARC-C, HellaSwag, PIQA, Winogrande, BoolQ, OBQA) as well as on reasoning via GSM8K. 
DLMs are evaluated using FastDLLM~\citep{wuFastdLLMTrainingfreeAcceleration2025} with 
a generation length of 1024 and one token decoded per diffusion step. Further details can be found in \autoref{app:details-eval}.

\paragraph{Compression.}
Pruning experiments were performed using WANDA \citep{sun2024simpleeffectivepruningapproach} and quantization experiments using GPTQ \citep{frantar2023gptqaccurateposttrainingquantization}.
Both methods used calibration samples drawn from the C4 dataset (128 for WANDA, 256 for GPTQ). Adding samples with 
masked tokens did not change the performance of the compression at all, so we opted not to mask samples for the DLM compression.
The described sparsity schedules \textit{earlier-is-sparser} (EIS) and \textit{deeper-is-sparser} (DIS) follow a simple linear allocation, using an $\epsilon$ of 0.08. 
This means for EIS applied to a $T$ layer network at sparsity $s$, we assign sparsity $s + 0.08\left(1 - \frac{2(t-1)}{T-1}\right)$ to layer $t$. 
Correspondingly, for DIS, the addition turns into a subtraction. 

\vspace{-6pt}
\subsection{Compression}
\vspace{-6pt}
\label{ssec:compression}

\begin{figure}[ht]
    \centering
    \includegraphics[width=\textwidth]{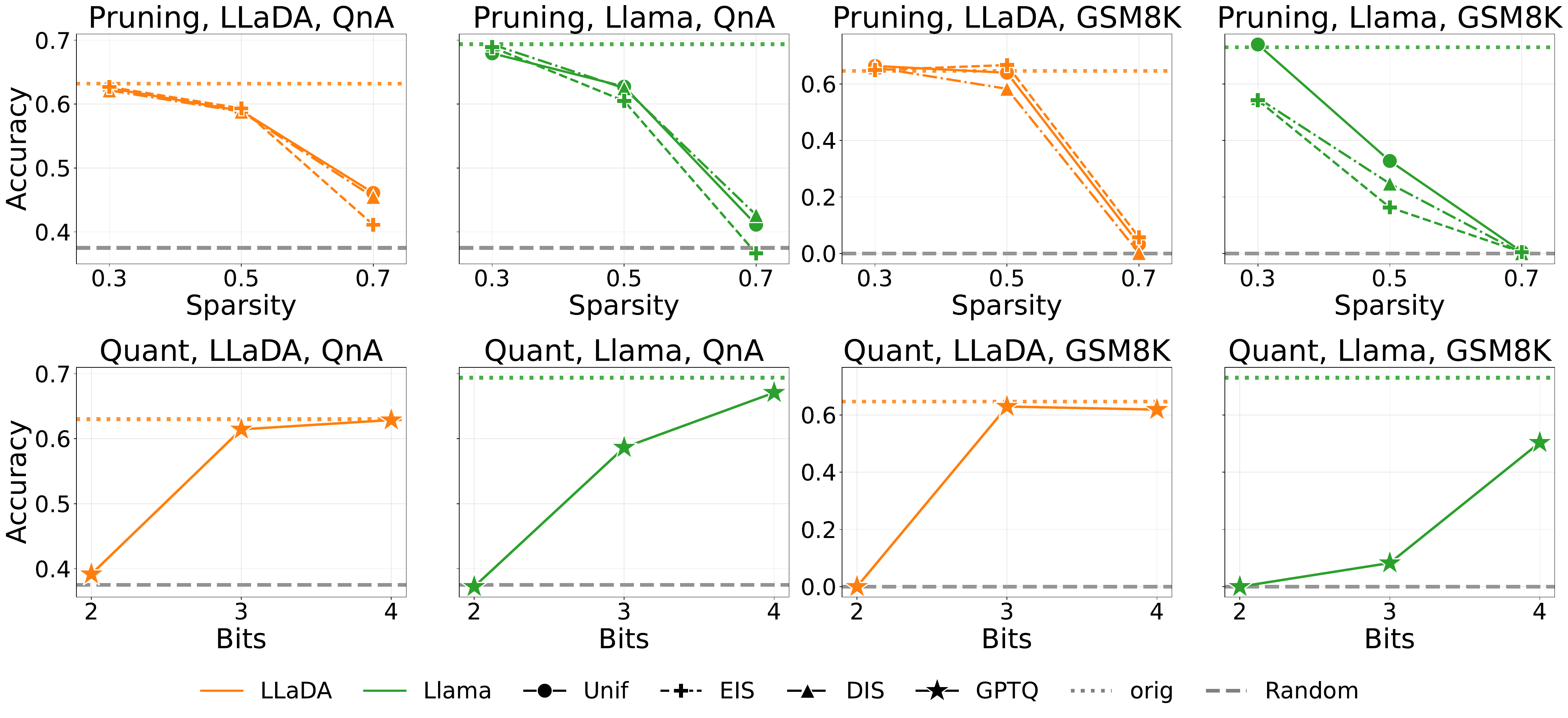}
    \caption{Compression performance for \llada and \llama: pruning across sparsity allocation strategies (top row, at $\epsilon=0.08$) and GPTQ quantization across bit-widths (bottom row), evaluated on the base-model commonsense QnA tasks (left column) and the instruct-model GSM8K task (right column).}\label{fig:summary-paper-combined}
\end{figure}

\paragraph{Pruning.} Our experiments in \autoref{fig:summary-paper-combined} (top row) show the average accuracy on QnA tasks (left) respectively GSM8K reasoning (right)
of \llada and \llama at three sparsity levels (0.3, 0.5, 0.7) and under three sparsity allocation strategies that result 
in the same total sparsity (uniform, earlier-is-sparser, deeper-is-sparser). 

These results reinforce the practicality of our observations made in \autoref{sec:dynamics}. We have two key observations to make in the results: 
\textbf{First}, \lladas starts out with a generally lower accuracy than \llamas, but is much more robust when compressed 
achieving almost double the accuracy of \llamas for GSM8K at 50\% sparsity. This ties into our observation 
that \llama contains many more similarities, introducing redundancy into the network and making the accuracy of
each single layer less important (as long as the super-outlier is protected, which is exactly the case under WANDA pruning). 
\textbf{Second,} the optimal sparsity allocation differs between \lladas and \llamas. For \llamas, pruning early layers more 
strongly (EIS) is always suboptimal and either uniform sparsity allocation or pruning deeper layers more strongly (DIS) are the optimal choice 
(depending on task). For \lladas, this flips, and DIS is usually the suboptimal choice, with either EIS or uniform performing the best. 
Only for sparsity 0.7 on QnA, DIS significantly outperforms EIS for \lladas, which we suspect might be due to the fact that the very high 
sparsity level starts to necessitate pruning of the weights connected to the super-outlier.

%

\paragraph{Quantization.}
As quantization is usually the more practical method of compressing a neural network, we have included
experiments on quantization as well in \autoref{fig:summary-paper-combined} (bottom row). However, as quantization has to be done to discrete
bit widths, we have not included an experiment on a layerwise bit-width allocation, which cannot be made with such a fine-grained
allocation as is done in the experiments on pruning.

Other than that, these experiments reinforce the same observation we have made for pruning above, namely that \lladas is inherently
more robust to compression than \llama, starting out with a worse base performance and surpassing the performance of \llamas at
3 bit for QnA (left figure) and already at 4 bit for GSM8K (right figure).

\vspace{-6pt}
\section{Supporting Evidence: A Controlled 160M Pair}
\vspace{-6pt}
\label{sec:smallscale}
The experiments in this section reinforce that our findings are not just a spurious quirk of \llada, but are inherent to the 
pre-training objective that is used in DLMs. To establish this, we perform pre-training experiments with a Pythia-160M \citep{Vaswanietal2017, Bidermanetal2023} architecture pre-trained on 100B tokens from FineWebEdu \citep{Penedoetal2024}. The resulting models \pyar, \pydlm are trained identically, 
solely varying in the training objective (cross-entropy either with or without masking). Further details are in \autoref{app:details-pretraining}. We perform the same evaluation as we did for \autoref{sec:experiments}, minus GSM8K due to the small model scale.

\vspace{-6pt}
\subsection{Results}
\vspace{-6pt}
\label{sec:smallscale-dynamics}
Our analysis of the small-scale models supports the following two observations from the base model:
(a) early layer representations in \pydlm are more similar to each other than in \pyar, and more similar to each other than
the deeper layer representations, and (b) \pydlm shows more robustness to compression than \pyar. Due to our controlled setting, we could thus ensure that these observations
are due to the DLM training objective and not due to architectural or pre-training hyperparameter choices.
We note that \pydlm is lacking the super-outlier: to observe weight and activation outliers, one has to train models to much larger scale (>6.7B indicated by \citet{dettmers2022gpt3}), which was out of the scope of this work. 

\paragraph{Layer Similarities.}
To more easily quantify how similar the layer representations in our two models are, we have averaged the per-layer similarities 
in a blockwise fashion: we have grouped the first 4 layers of the model together, as well as the last 4 layers and have averaged
over the resulting 16 inner-block layer-wise similarities. 
\autoref{tab:similarity-blocks-pythia-pair} shows that the same early-layer redundancy pattern of the large models starts to emerge 
in our pre-trained models: the early layer similarity is higher in \pydlm compared to \pyar, while it flips for the late-layer similarities.

\begin{table}
    \vspace{-12pt}
    \caption{Average cosine-similarities between early layers are higher for \pydlm. To better assess the similarity, we have grouped the first 4 layers and the last 4 layers together 
    and have averaged over their pairwise cosine similarities. }\label{tab:similarity-blocks-pythia-pair}
    \begin{center} \begin{tabular}{lcc}
\toprule
Model & Early-layer similarity & Deeper-layer similarity \\
\midrule
DLM-160M & 0.877 & 0.787 \\
AR-160M & 0.818 & 0.829 \\
\bottomrule
\end{tabular}

    \end{center}
    \vspace{-12pt}
\end{table}

\begin{figure}[ht]
    \begin{center}
        \includegraphics[width=0.8\textwidth]{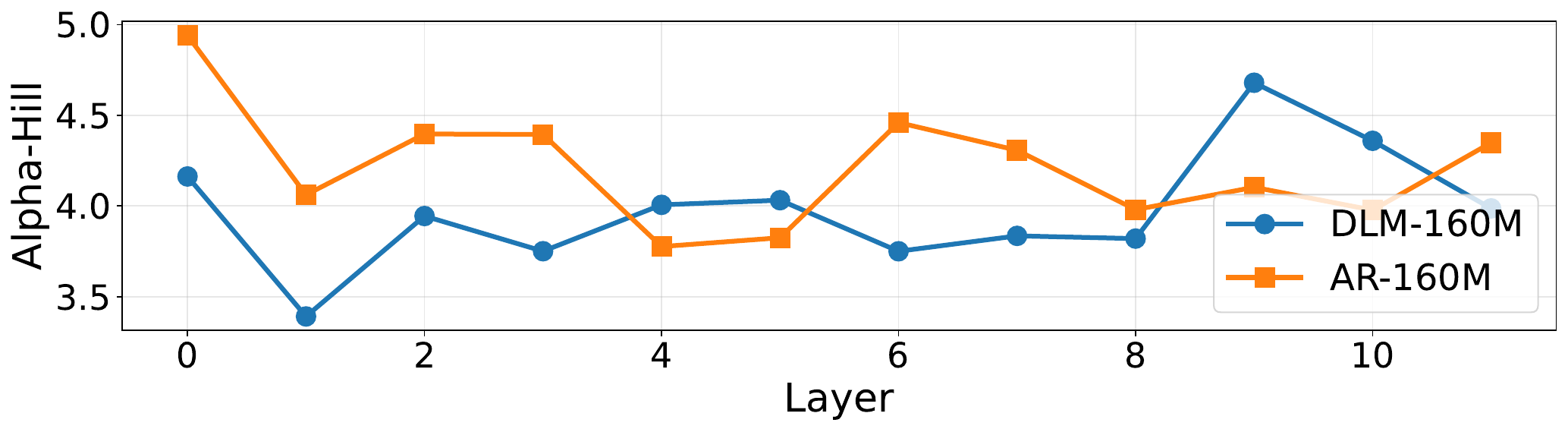}
    \end{center}
    \caption{Similar to the larger models, $\ahill$ is smaller for \pydlm compared to \pyar.}\label{fig:alpha-hill-pythia-pair}
    \vspace{-3pt}
\end{figure}

\paragraph{Heavy-tailed Signature.}
The same heavy-tailed signature appears in the 160M pair (\autoref{fig:alpha-hill-pythia-pair}): the controlled DLM shows a markedly lower $\ahill$ 
in early layers, while its AR twin does not. This behaviour normalizes in later layers, where both models show similar $\ahill$ values.

\begin{figure}[ht]
    \centering
    \includegraphics[width=\textwidth]{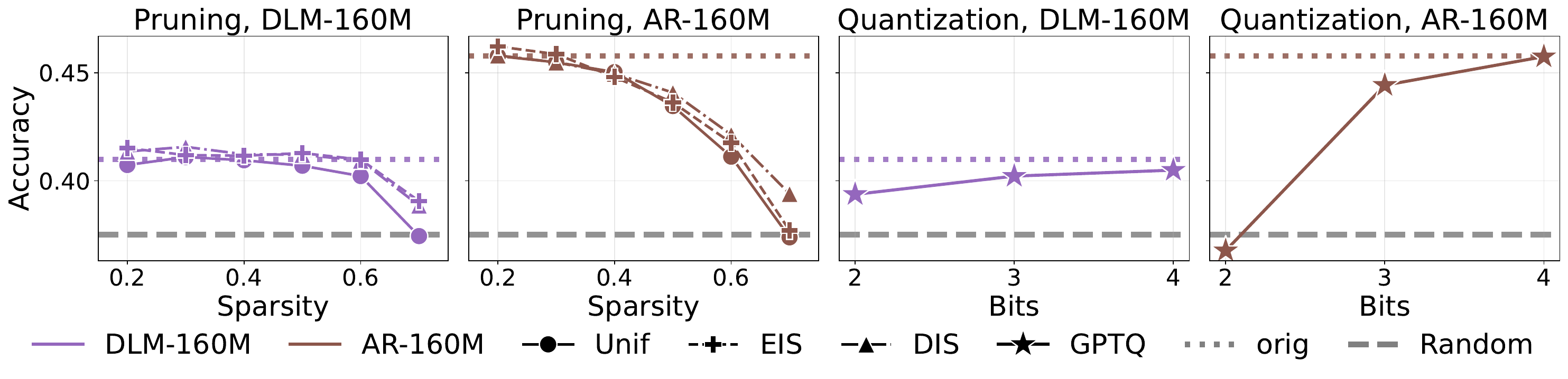}
    \caption{Robustness replication on the controlled 160M pair: pruning (left) and quantization (right). \pydlm loses less performance than \pyar under pruning, with the best sparsity allocation strategy inverted (EIS for the DLM, DIS for the AR). Under quantization, \pydlm starts below \pyar at full precision but surpasses it from 2 bits onward. Both results are in line with the behaviour of \llada versus \llama.}\label{fig:robustness-pythia-pair}
\end{figure}

\paragraph{Compression Results.}
\label{sec:smallscale-robustness}
In \autoref{fig:robustness-pythia-pair}, we present results for pruning (left) and quantization (right) of our 160M models. 
We have performed pruning and evaluation for all pre-training learning rates and selected the best performing model per 
sparsity level. Similar to the behaviour of \llada, \pydlm starts out with less performance than \pyar, but is more robust to compression. 
Here, the DLM model only supersedes its AR counterpart for the quantization experiment, while for pruning, they roughly
match at the highest sparsity level (which could be due to initially subpar performance of \pydlm). Also replicating the behaviour of \lladas versus \llamas, the best sparsity allocation strategy for \pydlm is consistently earlier-is-sparser,
while \pyar achieves higher accuracy using deeper-is-sparser (aside from low sparsity regimes up to 40\%, where all strategies perform similarly).
This matches the expectations from our similarity analysis in \autoref{tab:similarity-blocks-pythia-pair}. 

\vspace{-6pt}
\section{Related Work}
\vspace{-6pt}
\label{sec:background}

\paragraph{Outliers, Sinks, and Super-weights.}
A growing body of work shows that a small number of coordinates disproportionately
shape large language model behaviour. Attention sinks concentrate probability mass on
a handful of tokens \citep{xiao2024efficientstreaminglanguagemodels, gu2024attention};
massive activations identify a few token positions whose hidden states grow orders of
magnitude larger than the rest \citep{sunMassiveActivationsLarge2024}; super weights
\citep{yu2025superweightlargelanguage} and systematic outliers
\citep{an2025systematicoutlierslargelanguage} extend the picture to individual
parameters and feature dimensions whose removal disproportionately damages the model.
These phenomena are predominantly described along the \emph{token} axis: a small set
of positions dominates attention or activation norm. Collapsed layers describe the
\emph{channel-axis} counterpart: a single hidden dimension that dominates the
representation across all positions. In contrast to attention sinks, this phenomenon is absent in autoregressive models of comparable scale. 

\paragraph{Curse of Depth.} Multiple works analyze the tendency of AR models to have less well-trained 
deeper layers \citep{sun2025curse,gromov2025unreasonableineffectivenessdeeperlayers,Siji2026FromWT}, a phenomenon also termed the \emph{Curse of Depth}. Existing approaches
to mitigate this issue include depth-growing during training \citep{Kapl2025DoDM}, adding sparsity to the model \citep{muhtar2026doessparsitymitigatecurse} or mixing Pre- and Post-LN \citep{Li2024MixLNUT}. While our work reveals
that early layers in DLMs are more redundant and can be compressed more strongly than in ARs, this does 
not mean that the curse is broken:  if an improved training paradigm would make early-layers more informative and 
distinct,  DLMs still exhibit strong redundancies in later layers and could thus face similar issues.

\paragraph{Weight and Activation Dynamics in DLMs.}
A concurrent work by \cite{goel2026comparativeanalysislayerwiserepresentational} also uses the cosine-similarity of hidden-states to characterize layer-dynamics, but analyzes similarities between
immediately neighbored layers instead of long-range interactions, also revealing early-layer redundancies. The authors use this to propose a layer-skip method for inference,
which is different from our sparsity allocation.
In agreement with our work, the authors find that 
AR-finetuned DLMs retain most of the layer dynamics from 
initialization. Apart from that, our work uses cosine similarities in a markedly different way, which allows us to detect the super-outlier in \llada. Additionally, we provide
controlled AR-DLM twins and perform extensive compression experiments.
\citet{rulli2025attentionsinksdiffusionlanguage} characterise attention sinks in DLMs, which are token-level attention outliers. Our work is complementary, as we analyze channel-wide \emph{activation} outliers.

\paragraph{Compression of Diffusion Language Models.}
Several recent works compress masked diffusion language models. On the quantization
side, DLM-Quant \citep{xuDLLMQuantQuantizingDiffusionbased2025},
Quant-dLLM \citep{zhangQuantdLLMPostTrainingExtreme2025}, and
\citet{linQuantizationMeetsDLLMs2025} adapt post-training quantization pipelines
to the bidirectional, multi-step inference of DLMs. 
\citet{myrzakhan2026sinkawarepruningdiffusionlanguage} use insights on attention sinks in DLMs to improve pruning. 
These works treat compression as a methodological problem, where they adapt a known technique to a new architecture.
Our work is decidedly not a compression paper, but an analysis of activation dynamics whose findings carry practical implications that may 
be used in further compression research.

\vspace{-6pt}
\section{Discussion}
\vspace{-6pt}
\label{sec:discussion}
We have shown that DLMs exhibit activation dynamics that are qualitatively distinct from their AR counterparts. LLaDA-8B contains a single dominant activation channel that persists across all token positions and early layers; removing it causes total model collapse, a fragility with no parallel in AR models of comparable scale. This super-outlier drives early-layer redundancy through a mechanism we attribute to overtraining rather than undertraining, as evidenced by heavy-tailed weight spectra in early layers. A controlled 160M pre-training comparison supports that both the redundancy pattern and the compression robustness are products of the diffusion training objective and 
not architectural or pre-training artifacts. These findings carry direct practical consequences:
Allocating more sparsity to deeper layers, exactly what is useful for AR models, instead performs suboptimally on DLMs.

We envision that our findings inspire further research into DLM activation dynamics. 
The super-outlier in particular leaves many open questions, for example, which mechanisms of the DLM pre-training process produce it, and whether it is a harmful artifact or provides a useful role in information processing of DLMs.
Additionally, the redundancy of DLM layer representations might indicate that the pre-training process is not yet  optimized, similar to how redundant later layers in pre-LN AR models indicate suboptimal gradient flow. Ensuring that all layers are trained to produce distinct and informative
representations would produce models that are more capable at similar sizes.

\paragraph{Limitations.} Our findings are mostly focussed on a single DLM-AR pair, namely \llada and \llama. While the behaviour of \llama 
is consistent with what is reported in the AR model literature, our findings for \llada are novel. 
To our knowledge, \llada is the only large-scale DLM that is completely pre-trained from scratch, as 
the current literature is mostly focussed on creating DLMs by fine-tuning them from AR model weights 
\citep{bie2025llada20scalingdiffusionlanguage,bie2026llada21speedingtextdiffusion,yu2026introspectivediffusionlanguagemodels}.
During fine-tuning, these models' weights stay close to their original, therefore models 
such as DREAM-7B show activation dynamics that are very close
to their AR counterpart, see \autoref{fig:similarity-plots}. While supporting  our other findings, the 160M pre-training models did not show the super-outlier as \llada, which was expected given the small scale of the model. To make our results more widely applicable, 
a large-scale DLM pre-training study could be implemented, pre-training DLMs on multiple parameter scales (e.g. 350M, 1B, 3B, 8B, 16B). This task unfortunately requires large amounts of GPU resources to run.

\bibliographystyle{abbrvnat}
\bibliography{aconzelmann}

\appendix 
\section{Experimental details}
\label{app:details}
\subsection{Evaluation of Language Models}
\label{app:details-eval}
Models are evaluated on the following question-answering tasks:
ARC-Challenge~\citet{clark2018thinksolvedquestionanswering},
HellaSwag~\citet{zellers2019hellaswagmachinereallyfinish},
PIQA~\citet{bisk2019piqareasoningphysicalcommonsense}
WinoGrande~ \citet{sakaguchi2019winograndeadversarialwinogradschema},
BoolQ~\citet{clark2019boolqexploringsurprisingdifficulty},
OpenbookQA~\citet{OpenBookQA2018}. Additionally, we evaluate on reasoning via GSM8K~\citep{cobbe2021gsm8k}. 
We use 25-shot for ARC-Challenge, 10-shot for HellaSwag, 5-shot for WinoGrande and GSM8K, and 0-shot for BoolQ, OpenBookQA, and PIQA.
We used base models for QnA, and corresponding instruction-finetuned variants for GSM8K.  

For the DLMs, we used an adapted version FastDLLM~\citep{wuFastdLLMTrainingfreeAcceleration2025} with single KV-cache, but not parallel decoding. The block-length was set to 32 and the generation length to 1024, with 1024 decoding steps (so 1 token per decoding step) and low confidence remasking.  
For the AR models, we use vLLM \citep{kwon2023efficient} for inference acceleration together with LM-Eval \citep{eval-harness}.

All evaluations are done on a single H100 GPU. Evaluations take around 6 hours for DLMs and 20 minutes for AR models for both task sets.

\subsection{Small-scale pre-training}
\label{app:details-pretraining}
\paragraph{Autoregressive Model.} 
We used \citet{ajroldi2024plainlm} to pretrain Pythia-160M parameter transformer models \citep{Vaswanietal2017, Bidermanetal2023} on causal language modeling, with 100B tokens of FineWebEdu \citep{Penedoetal2024} on 8$\times$A100-80GB GPUs. We use sequence length 2048 and a batch size of 0.5M tokens, cross-entropy loss, Adam \citep{KingmaBa2015} with decoupled weight decay \citep{LoshchilovHutter2019} of 0.1, gradient clipping of 1, and $(\beta_1,\beta_2)=(0.9, 0.95)$. We use Warm up-Stable-Decay \citep{wen2024wsd} to schedule the learning rates, warm up of 1900 steps ($1\%$) and decay to 0 \citep{bergsma2025straightzerolinearlydecaying} of $10\%$ of token budget.
We perform three independent runs for learning rates $\{3\!\times\!10^{-3}, 1\!\times\!10^{-3}, 3\!\times\!10^{-4}\}$.

\paragraph{Diffusion Language Model.} 
We adapt this pipeline into a Masked Discrete Diffusion Language Model 
\citep[MDLM;][]{Sahooetal2024, Louetal2024} trainer with four modifications: 
(i) a bidirectional attention patch on the GPTNeoX backbone, 
(ii) a forward absorbing-state corruption step, 
(iii) an importance-weighted cross-entropy loss applied only at corrupted positions, 
and (iv) the reuse of an unused vocabulary slot as the $[\text{MASK}]$ token. 
All other hyperparameters, including data order, are identical to the AR trainer.

\section{Further experiment plots}
\subsection{Activations and similarities without masking}
\label{app:no-mask}
We replicate a subset of our experiments while calculating activations in DLMs without 
masking tokens to showcase that our findings are robust over diffusion steps. 

\begin{figure}[h]
    \centering
    \includegraphics[width=0.7\textwidth]{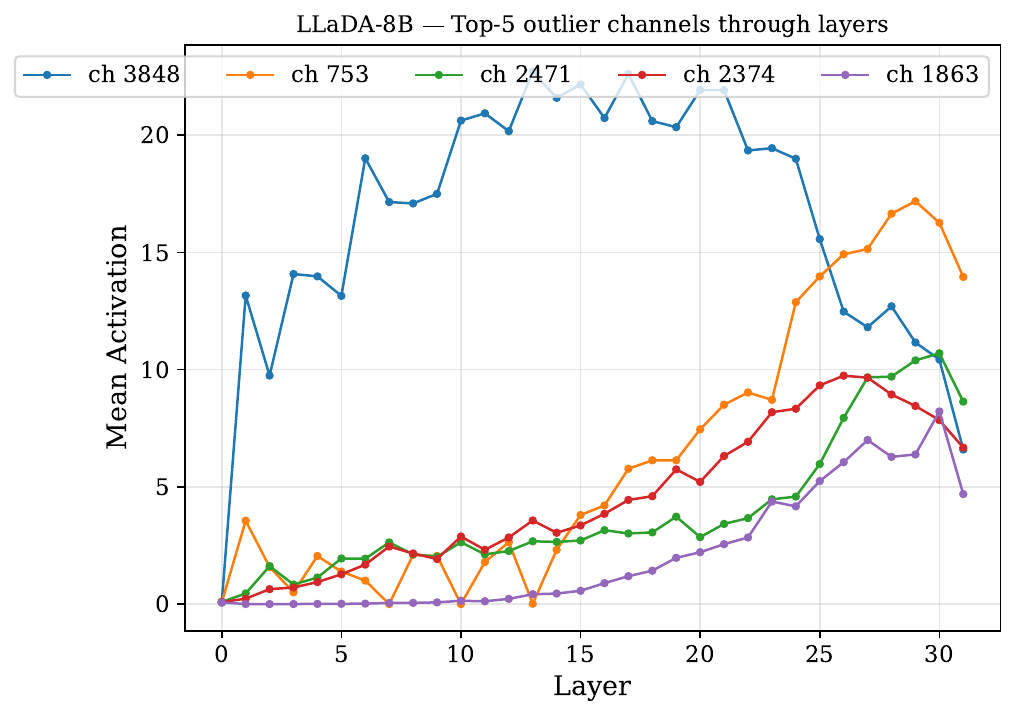}
    \caption{Similar to \autoref{fig:llada-top5}, but without including masked sequences.}
\end{figure}

\begin{figure}
    \begin{center}
        \includegraphics[width=0.95\textwidth]{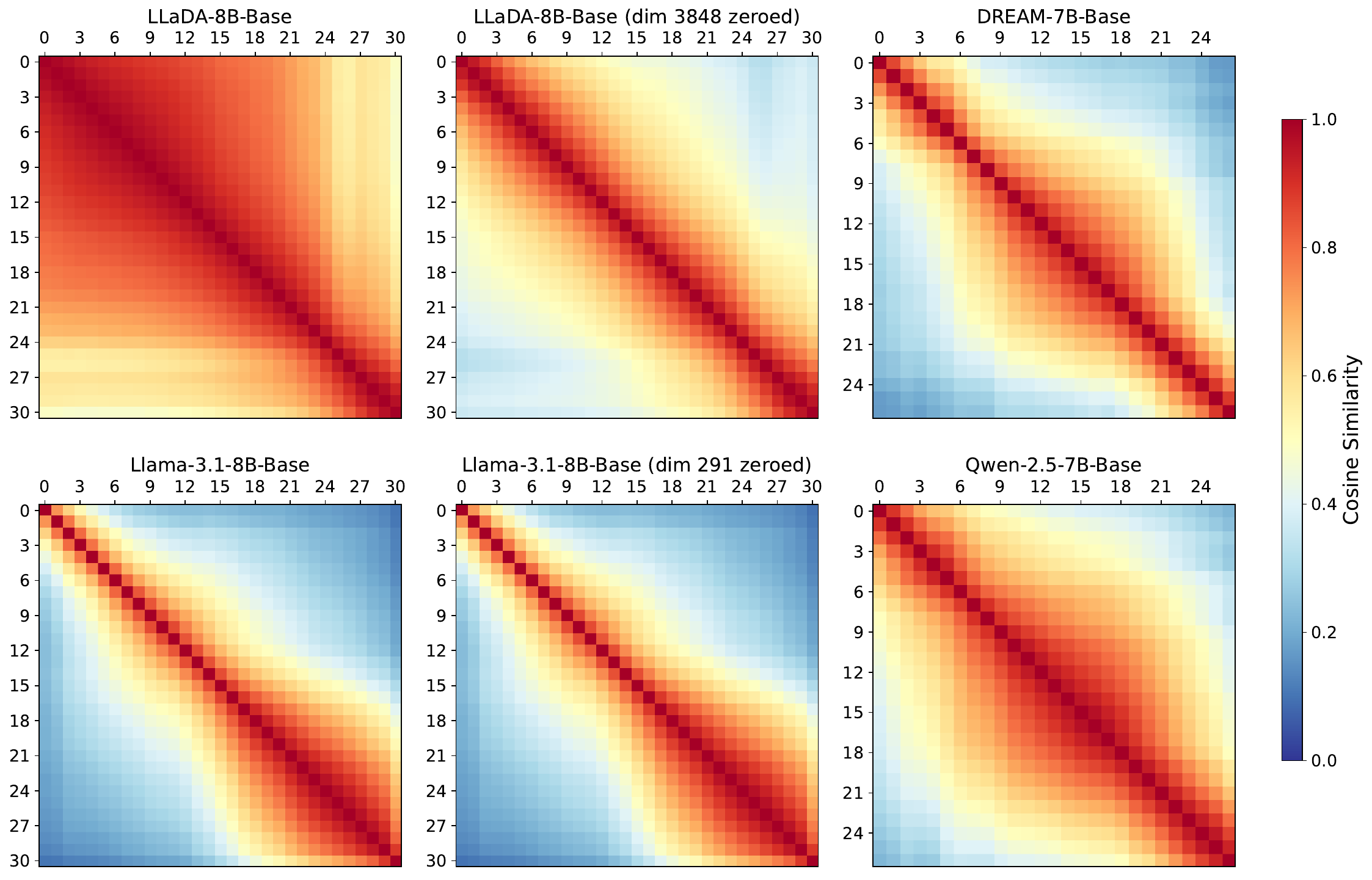}
    \end{center}
    \caption{Similar to \autoref{fig:similarity-plots}, but without including masked sequences.}\label{fig:similarity-plots-nomask}
\end{figure}

\FloatBarrier
\subsection{Extended activation plots}
\begin{figure}[h]
    \centering
    \includegraphics[width=0.7\textwidth]{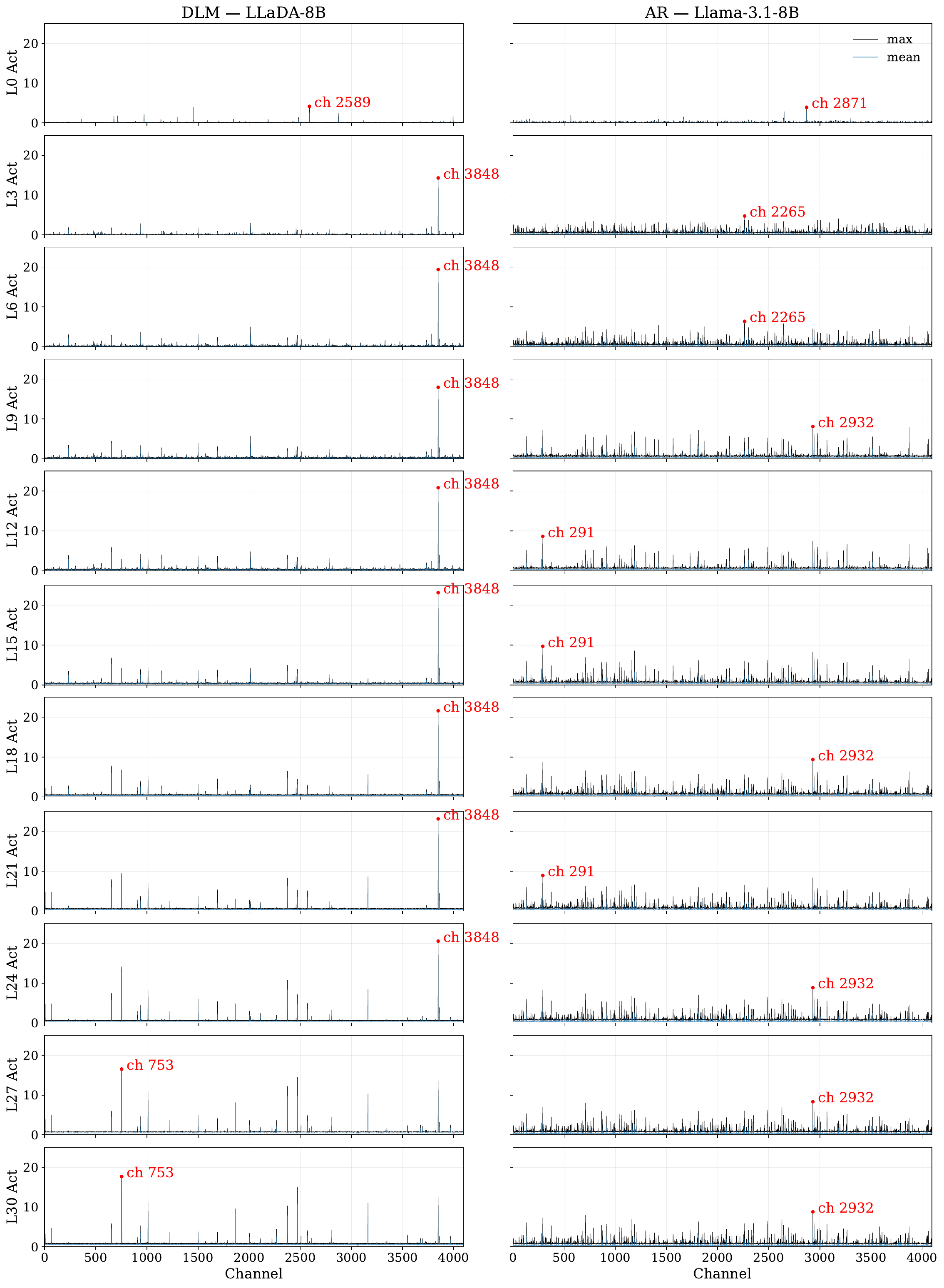}
    \caption{Extended version of \autoref{fig:sparkline}.}
\end{figure}

\FloatBarrier
\newpage
\subsection{Channel activation over Diffusion Steps}

\begin{figure}[h]
    \centering
    \includegraphics[width=0.9\textwidth]{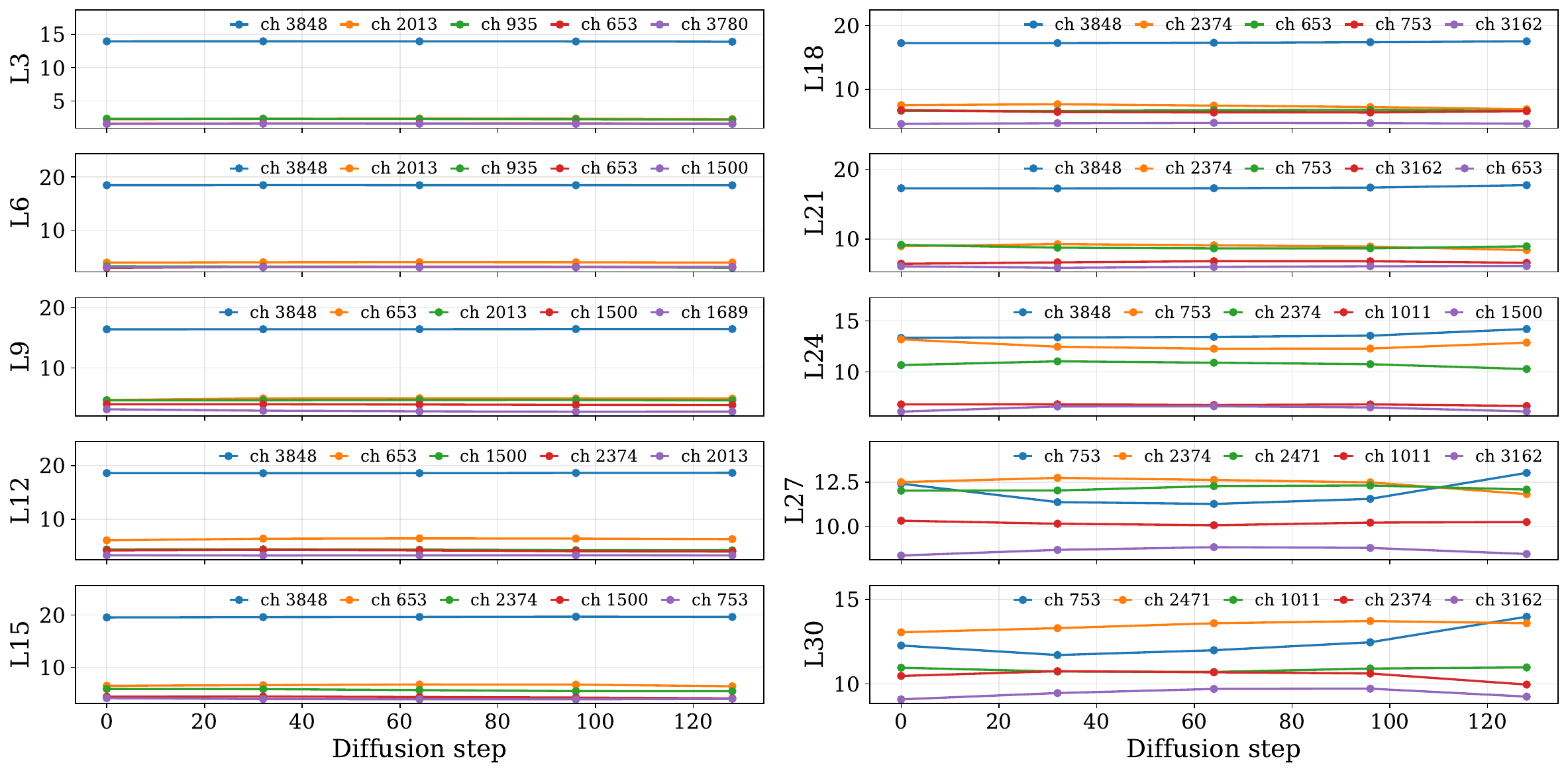}
    \caption{Channel magnitude mean of the top-5 largest (by mean) channels in \llada, over different diffusion steps. The channel
    magnitudes for early-mid layers barely change over the diffusion step. }
\end{figure}
%

\FloatBarrier
\end{document}